\documentclass[10pt,twocolumn,letterpaper]{article}

\usepackage{threeparttable}

\usepackage[pagenumbers]{iccv} 

\definecolor{iccvblue}{rgb}{0.21,0.49,0.74}
\usepackage[pagebackref,breaklinks,colorlinks,allcolors=iccvblue]{hyperref}

\def\paperID{*****} 
\def\confName{ICCV}
\def\confYear{2025}

\title{A Signer-Invariant Conformer and Multi-Scale Fusion Transformer for Continuous Sign Language Recognition}

\author{
Md Rezwanul Haque\textsuperscript{1 *},
Md. Milon Islam\textsuperscript{1 *},
S M Taslim Uddin Raju\textsuperscript{1},
Fakhri Karray\textsuperscript{1,2} \\
\textsuperscript{1}Department of Electrical and Computer Engineering, University of Waterloo\\
\textsuperscript{2}Department of Machine Learning, Mohamed bin Zayed University of
Artificial Intelligence \\
{\tt\small \textsuperscript{1}\{rezwan, milonislam, smturaju, karray\}@uwaterloo.ca, \textsuperscript{2}fakhri.karray@mbzuai.ac.ae}
}

\begin{document}
\maketitle

\begingroup
\renewcommand\thefootnote{}
\vspace{-1em} 

\footnotetext{
\textsuperscript{*}Equal contribution. 
Correspondence to: Md Rezwanul Haque \texttt{<rezwan@uwaterloo.ca>}.
}

\footnotetext{
\textsuperscript{©}\textit{Proceedings of the IEEE/CVF International Conference on Computer Vision (ICCV), Honolulu, Hawaii, USA. 1st MSLR Workshop 2025. Copyright 2025 by the author(s).}
}

\endgroup

\begin{abstract}
Continuous Sign Language Recognition (CSLR) faces multiple challenges, including significant inter-signer variability and poor generalization to novel sentence structures. Traditional solutions frequently fail to handle these issues efficiently. For overcoming these constraints, we propose a dual-architecture framework. For the Signer-Independent (SI) challenge, we propose a Signer-Invariant Conformer that combines convolutions with multi-head self-attention to learn robust, signer-agnostic representations from pose-based skeletal keypoints. For the Unseen-Sentences (US) task, we designed a Multi-Scale Fusion Transformer with a novel dual-path temporal encoder that captures both fine-grained posture dynamics, enabling the model's ability to comprehend novel grammatical compositions. Experiments on the challenging Isharah-1000 dataset establish a new standard for both CSLR benchmarks. The proposed conformer architecture achieves a Word Error Rate (WER) of 13.07\% on the SI challenge, a reduction of 13.53\% from the state-of-the-art. On the US task, the transformer model scores a WER of 47.78\%, surpassing previous work. In the SignEval 2025 CSLR challenge, our team placed $2^{nd}$ in the US task and $4^{th}$ in the SI task, demonstrating the performance of these models. The findings validate our key hypothesis: that developing task-specific networks designed for the particular challenges of CSLR leads to considerable performance improvements and establishes a new baseline for further research. The source code is available at: \url{https://github.com/rezwanh001/MSLR-Pose86K-CSLR-Isharah}.
\end{abstract}    
\section{Introduction}
\label{sec:intro}

Sign language is a rich visual-gestural language that plays a vital role in communication for deaf and hard-of-hearing individuals \cite{desisto2023survey}. According to the World Health Organization, over 5.5\% of the global population is affected by hearing loss, making accessible communication technologies increasingly important \cite{who}. Sign languages convey meaning through a combination of manual features such as hand shapes and movements, and non-manual cues such as facial expressions and body posture \cite{el2022comprehensive}. Continuous sign language recognition has emerged as a crucial task in this context, aiming to convert sequences of sign gestures into textual representations \cite{alyami2024reviewing}. Compared to Isolated Sign Language Recognition (ISLR), CSLR poses greater challenges due to the absence of clear boundaries between signs and the effects of co-articulation \cite{aloysius2020understanding}. Furthermore, effective CSLR systems must generalize across diverse signing styles and signer-specific variations, requiring robust modeling of temporal and contextual dependencies \cite{luqman2019automatic}.

Continuous sign language recognition remains a challenging task due to practical limitations in data and the inherent complexity of sign language structure. Most existing datasets focus on isolated signs, as they are easier to collect and annotate, while sentence-level datasets require skilled signers and detailed gloss annotations, making them costly and time-consuming \cite{sidig2021karsl, li2020word}. Furthermore, many existing datasets are recorded in highly controlled environments, reducing their effectiveness for real-world deployment \cite{adaloglou2021comprehensive, kagirov2020theruslan}. Recently, the Isharah dataset has addressed some of these issues by collecting diverse, unconstrained videos, but large-scale, richly annotated CSLR data \cite{adaloglou2021comprehensive}. On the other hand, temporal dependencies in continuous signing remain challenging, as co-articulation and unclear sign boundaries affect frame-to-gloss alignment \cite{athira2022signer, aloysius2020understanding}. To address this, weakly supervised methods such as Connectionist Temporal Classification (CTC) and Dynamic Time Warping (DTW) are often used, though their performance is limited in generalization tasks \cite{aloysius2020understanding}. As a result, techniques such as CTC and DTW are commonly applied, although they struggle in signer-independent and unseen-sentence scenarios \cite{aloysius2020understanding}.

Recent studies have explored deep learning architectures that separately model spatial and temporal features to address the challenges of CSLR.  Spatial information is frequently captured using 2D/3D Convolutional Neural Networks (CNNs), graph convolutional networks, or vision transformers to process hand movements and non-manual cues across frames \cite{li2022multi, cui2023spatial}. For temporal modeling, Recurrent Neural Networks (RNNs) and temporal convolutional networks are widely used, yet these methods often struggle to capture the co-articulated, context-dependent nature of continuous signing \cite{aloysius2020understanding}. Moreover, most existing models rely on benchmark datasets such as RWTH-PHOENIX-Weather \cite{forster2014extensions} and CSL \cite{huang2018video}, which limits vocabulary range and generalization to real-world tasks. Although attention-based models such as transformers offer improved temporal modeling, challenges like signer variation and subtle motion patterns still exist. These issues underline the need for more unified frameworks that can capture both local and global temporal dynamics and perform reliably across diverse signing conditions.

The key contributions of this research are outlined as follows:

\begin{itemize}

\item We propose a novel architecture, Signer-Invariant Conformer, which combines convolutional and self-attention methods. This demonstrates the ability to learn signer-agnostic representations for the SI CSLR issue.

\item We introduce a Multi-Scale Fusion Transformer that uses a joint attention mechanism and a novel dual-path temporal encoder. This network aims to enhance language generalization by capturing fine-grained pose dynamics for the US challenge.

\item We perform extensive tests on the challenging Isharah-1000 dataset, achieving new outstanding scores on both SI and US benchmarks. Our findings provide a new baseline and provide the effectiveness of developing task-specific networks for CSLR.
\end{itemize}

The remaining sections of this paper are organized as follows. Section \ref{sec:related works} analyzes previous CSLR research, keeping our research within the existing domain of datasets and methods. Section \ref{sec:method} describes our methods in depth, including pose-based data representation and the architectures of both Signer-Invariant Conformer and Multi-Scale Fusion Transformer. In section \ref{sec:experiment}, we detail our experimental setup and provide a thorough analysis of our results on the Isharah benchmark, comparing our networks against State-of-the-Art (SOTA) methods. Lastly, section \ref{sec:conclusion} summarizes our findings, and outlines potential future research directions.

\section{Related Work}
\label{sec:related works}

\subsection{Benchmark Datasets in CSLR}
Alyami et al. \cite{alyami2025isharah} introduced Isharah, a large-scale multi-scene dataset for CSLR that was recorded entirely in unconstrained real-world environments using smartphone cameras. The dataset consisted of 30,000 videos encompassing over 2,000 unique Saudi sign language sentences performed by 18 fluent signers, with diverse backgrounds and signer variability. It included detailed gloss-level and spoken language annotations supporting CSLR and Sign Language Translation (SLT) tasks. Benchmarking seven CSLR models and two SLT methods revealed significant challenges in SI and US scenarios, with WER as low as 27.4\% for SI recognition and BLEU-4 scores up to 66.1 for gloss-based SLT.  Mukushev et al. \cite{mukushev2022fluentsigners} presented FluentSigners-50, a large-scale signer independent dataset for Kazakh-Russian sign language to benchmark CSLR and translation. It contains 43,250 videos of 173 sentences performed by 50 signers of different ages, genders, and hearing statuses recorded in real-life using smartphones and webcams. The dataset features high linguistic and signer variability, including phonetic, lexical, and syntactic differences, making it more realistic than previous controlled datasets. Baseline tests with stochastic CSLR and Temporal Semantic Pyramid Network (TSPNet) models on three splits showed WER between 24.9 and 52.0 and BLEU-4 scores from 16.0 to 2.0. FluentSigners-50 is publicly available to support robust sign language recognition and translation research.

The RWTH-PHOENIX-Weather 2014 corpus \cite{forster2014extensions} is a large-vocabulary dataset for German Sign Language (DGS), considered as a foundational benchmark in this area. This dataset is distinguished by its source, which consists of real-world weather forecasts shown on German public television and interpreted live by DGS. The 2014 version greatly expanded on its predecessor, with about 6,800 sentences signed by nine independent signers. A fundamental challenge of this corpus is its continuous and naturalistic signing, which is done in real-time, resulting in co-articulation and fast signing speeds. The dataset contains detailed annotations, such as glosses, spoken German transcripts, and spatial annotations for hand and face positions, making it a major resource for sign language recognition and translation research. The SIGNUM database \cite{von2010signum} was one of the first large-scale efforts to address the issue of signer-independent CSLR. It includes 780 unique sentences in DGS, each performed by 25 different native signers, comprising around 33,000 video sequences. The corpus was captured in a controlled laboratory environment with consistent lighting and a blue background to allow for effective feature extraction. This database has a vocabulary of 450 fundamental signs, provides an extensive repository of interpersonal variation and remains a key benchmark for developing and assessing models capable of generalizing over a different group of signers.

\subsection{Architectural Approaches for CSLR}

Huang et al. \cite{huang2018video} proposed the  Hierarchical Attention Network with Latent Space (LS-HAN) framework for CSLR that removed the need for temporal segmentation by using a hierarchical attention network combined with a latent space model. The method utilized a two-stream 3D CNN to extract global and local features and jointly optimized recognition and video-sentence relevance losses for end-to-end sentence prediction. Evaluations on a large Chinese sign language dataset (25,000 videos) and the RWTH-PHOENIX-Weather dataset showed that LS-HAN achieved accuracies up to 82.7\%, outperforming SOTA methods such as LSTM-E and DTW-HMM. This demonstrated the framework’s effectiveness and robustness compared to segmentation-dependent approaches. In another study, Zhou et al. \cite{zhou2020spatial} proposed the Spatial-Temporal Multi-Cue (STMC) network for CSLR, combining spatial multi-cue features with a temporal module modeling intra- and inter-cue correlations. The joint optimization with CTC enabled end-to-end training of hands, face, and pose expressions without external tools. The STMC network obtained SOTA WER of 20.7\% on PHOENIX-2014, 2.1\% on CSL (split I), and 21.0\% on PHOENIX-2014-T. This method improved recognition accuracy while maintaining computational efficiency.

Many studies now focus on using a signer's pose data as the main input for CSLR. This approach is popular because it is computationally efficient and is not affected by distracting backgrounds. The Spatial-Temporal Graph Convolutional Network (ST-GCN) \cite{yan2018spatial} treats the skeleton as a connected graph to understand the structure of the human body. Another method involves using transformer-based models, for example, Cui et al. \cite{li2022multi} used transformer-based models to capture long-range temporal relationships between pose sequences.

In another research, Zhang et al. \cite{10.1145/3377553} proposed WiSign, a continuous American sign language recognition system using commercial Wi-Fi devices. WiSign leveraged Channel State Information (CSI) signals, utilizing a five-layer deep belief network combined with Hidden Markov Model (HMM) and an N-gram language model for sign sentence segmentation as well as classification. The system scored up to 92\% accuracy with personalized models and 69\% accuracy with a general model across 30 users. The system achieved over 80\% accuracy at distances up to 3 meters while operating with sampling rates of at least 1000 Hz. Furthermore, Camgoz et al. \cite{cihan2017subunets} introduced SubUNets, an end-to-end deep learning method for CSLR that explicitly modeled subunits like hand shapes. This approach obtained SOTA hand shape recognition with 80.3\% Top-1 accuracy, representing a 30\% improvement over previous methods, without requiring separate alignment. Combining SubUNets for hand shape and full-frame data enhanced sentence-level recognition, scoring a WER of 42.1\%. Cui et al. \cite{cui2019deep} developed a deep neural network framework for CSLR that combines CNNs with temporal fusion layers and bi-directional RNNs. An iterative training process was introduced to refine alignment between video segments and gloss labels, enhancing recognition accuracy. The framework integrated RGB images and optical flow for multimodal fusion to capture both appearance and motion information. Evaluations on the RWTH-PHOENIX-Weather 2014 and SIGNUM datasets demonstrated SOTA performance, achieving more than 15\% relative improvement in WER. Multimodal fusion and iterative training contributed significantly to the enhanced results.
\section{Method}
\label{sec:method}

Continuous sign language recognition poses two major challenges: recognizing high inter-signer heterogeneity and generalizing to unknown syntactic patterns. Our approach addresses these challenges by using pose-based skeletal data, which provides an abstract representation of the signing motion. We divide our methodology into two specific pose-based architectures designed for the SI and US tasks. For the SI task, the proposed framework employs a conformer architecture. Its inherent combination of local convolutional processing and global self-attention is suitable for learning effective, signer-agnostic features that are robust to different signing styles. The approach for the US task utilizes a transformer-based architecture with a novel dual-path temporal encoder. By combining multi-scale features, this network gives more comprehensive input to the transformer, improving its ability for modeling the long-range dependencies required for interpreting grammatical compositions. The details of each architecture are described in the following sections.

\begin{figure*}[t]

\centerline{\includegraphics[trim={0.5cm 1.5cm 1.5cm 1.5cm}, scale=.56, angle=0]{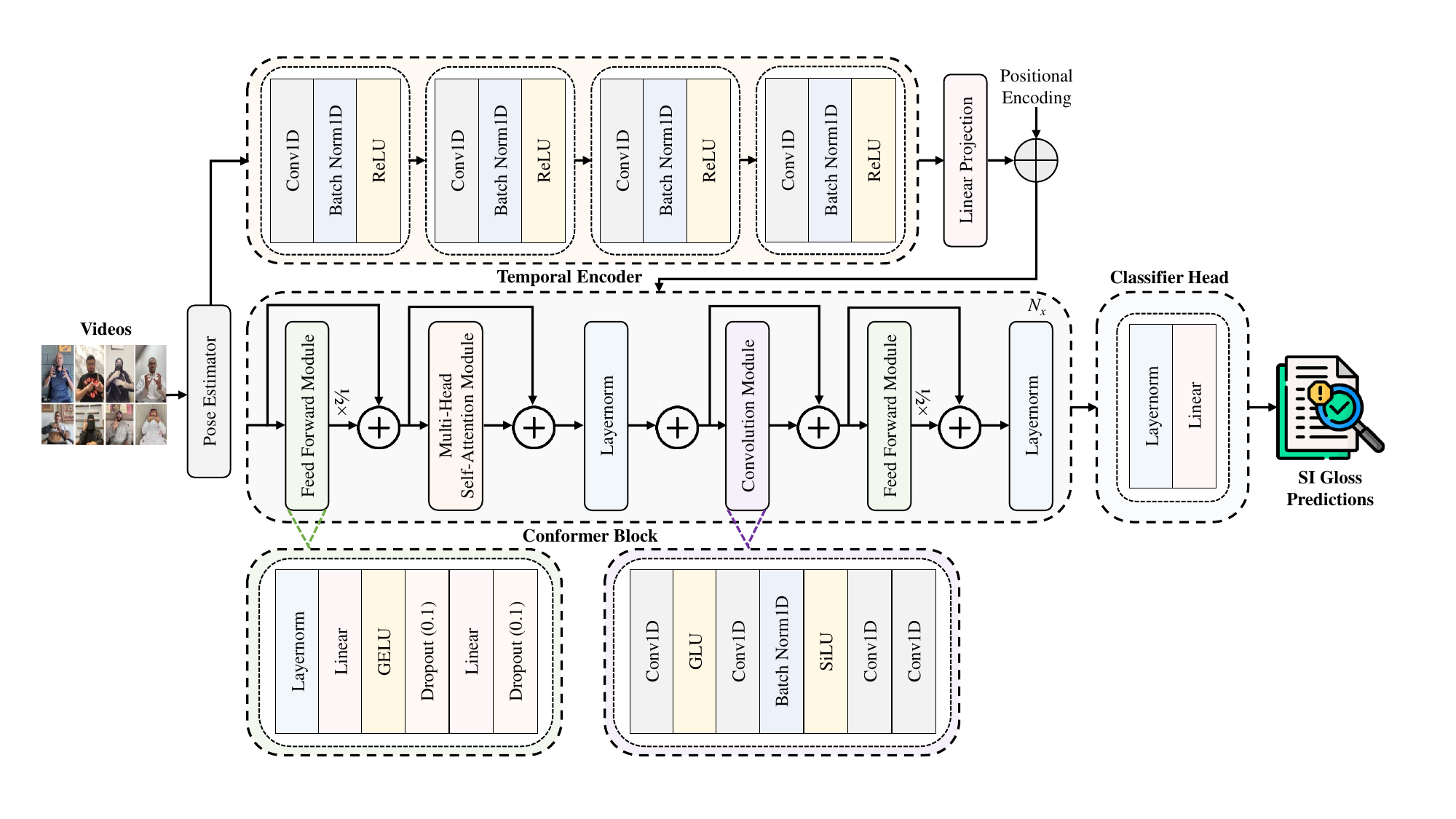}} 

\caption{\textbf{Signer-Invariant Conformer:} our proposed architecture for signer-independent CSLR begins by extracting pose keypoints from video frames. An initial temporal encoder, composed of convolutional layers, learns local features from this pose sequence. The core of the model consists of conformer blocks that capture global context with multi-head self-attention and extract local patterns using convolution. Positional encodings are exploited to provide the model with sequence order information. Finally, a linear classifier head analyzes the sequence representation to generate sign gloss predictions.}
\label{fig:Challenge 12}
\end{figure*}

\subsection{Data Representation and Preprocessing}
A pose-based approach is used that converts raw pixel data into a more comprehensive and computationally efficient skeletal representation. For each input video, a keypoint extraction framework is utilized to derive a comprehensive set of 2D landmarks corresponding to the pose, hands, and face. This generates a sequence of keypoint vectors, \(X = \{{x}_1, {x}_2, \dots, {x}_T\}\), where \(T\) is the number of frames and each \(\mathbf{x}_t \in \mathbb{R}^{K \times 2}\) represents the (x, y) coordinates of \(K=86\) landmarks.

To ensure robustness against variations in signer distance and position relative to the camera, a normalization technique is applied. The coordinates are normalized relative to a bounding box defined by the signer's torso, ensuring a consistent scale and reference frame across all samples. Missing keypoints, generated by occlusion or motion blur, are imputed using linear interpolation from adjacent frames. The final preprocessed input for the models is represented as a flattened tensor \({X}' \in \mathbb{R}^{T \times D}\), where \(D=K \times 2\) denotes the total number of keypoint coordinates.

\subsection{Conformer for SI Recognition}
The architecture of our SI network, shown in Fig. \ref{fig:Challenge 12}, is designed for generating a representation that is independent of signer characteristics. It leverages a conformer encoder that has demonstrated remarkable success in speech recognition by effectively capturing both local and global dependencies in sequential data \cite{gulati2020conformer}.

\subsubsection{Temporal Encoder}
The normalized keypoint sequence \({X}'\) is initially passed through a shallow temporal encoder. This module consists of a series of 1D convolutional layers with batch normalization and ReLU activations. Its primary function is to perform an initial feature projection and capture short-range local correlations between keypoints within a small temporal window. This step transforms the raw coordinate data into a more discriminative feature space, preparing it for the more complex sequential modeling.

\subsubsection{Conformer-based Sequential Encoder}
The core of our SI model is a stack of conformer blocks. Before entering the stack, the feature sequence is augmented with sinusoidal positional encodings to provide the model with information about the order of the signs. A conformer block integrates multi-head self-attention, depthwise convolutions, and feed-forward layers.

The Multi-Head Self-Attention (MHSA) mechanism captures the global, long-range dependencies across the entire sign sequence. For an input sequence \(Z\), it is calculated as in (\ref{eq:mhsa}).
\begin{equation}
\label{eq:mhsa}
 \text{MHSA}(Z) = F_{concat}(\text{head}_1, \dots, \text{head}_h)W^O 
\end{equation}
where each head is defined in (\ref{eq:head_i}).
\begin{multline}
\text{head}_i = F_{Attention}(ZW_i^Q, ZW_i^K, ZW_i^V) \\
= F_{softmax}\left(\frac{(ZW_i^Q)(ZW_i^K)^T}{\sqrt{d_k}}\right)ZW_i^V
\label{eq:head_i}
\end{multline}
This allows the model to weigh the importance of different signs in the sequence to understand co-articulation effects \cite{athira2022signer}.

Following the attention module, a convolution module comprising a 1D depthwise convolution layer with a Gated Linear Unit (GLU) activation is used. This module performs well at capturing local, relative positional information and fine-grained gestural patterns that the self-attention mechanism may overlook. The outputs of the MHSA and convolution modules are integrated within the block via residual connections and layer normalization, creating a robust hybrid representation.

\begin{figure*}[t]

\centerline{\includegraphics[trim={0.5cm 3.5cm 0.5cm 1.75cm}, scale=.55, angle=0]{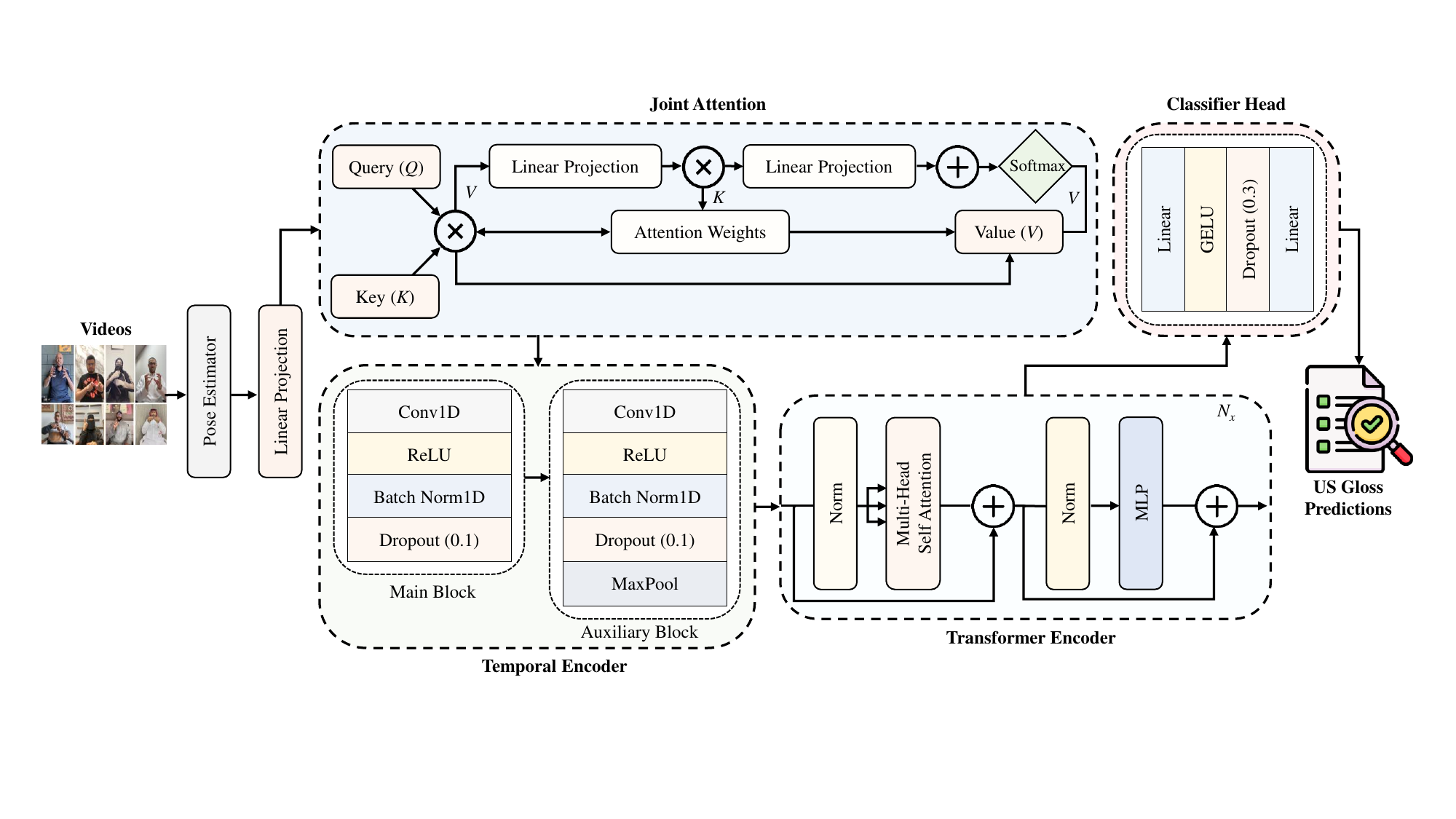}} 

\caption{\textbf{Multi-Scale Fusion Transformer:} an overview of the proposed architecture for the unseen sentences CSLR task. The network first uses a pose estimator to retrieve keypoint data. A joint attention module is then immediately applied to weigh the importance of features at the frame level. The generated features are then processed by a temporal encoder with a dual-path design: a main block records fine-grained temporal dynamics, and an auxiliary block uses max-pooling to learn downsampled representations. The outputs of both blocks are combined to provide a comprehensive feature set. The output sequences subsequently analyzed by a transformer encoder, which models the sequence's long-range relationships. The output feature vectors are fed into a classifier head, which generates the US gloss predictions.}
\label{fig:Challenge 11}
\end{figure*}

\subsubsection{Classifier Head}
The output of the final conformer blocks is a sequence of contextualized feature vectors. This sequence is processed by a classifier head, which consists of a layer normalization step followed by a linear layer that projects the feature vectors into the gloss vocabulary space of size \(|V_g|\). The model is trained using the CTC loss function. The CTC is suitable for CSLR because it allows for end-to-end training without requiring explicit, frame-level alignment between the video and the gloss sequence, by aggregating the probabilities of all possible alignments that result in the target gloss sequence.

\subsection{Multi-Scale Fusion Transformer for US Recognition}
To tackle the US task, a model should have linguistic generalization capabilities. Our proposed architecture, illustrated in Fig. \ref{fig:Challenge 11}, is designed to achieve this through a multi-stage feature extraction process. The pipeline begins with a joint attention mechanism that immediately weighs the importance of features at the frame level, enabling the model to focus on the most salient gestural information. These features are then processed by a novel dual-path temporal encoder. This design simultaneously captures fine-grained temporal dynamics in one path while learning downsampled representations in another branch. By fusing these complementary multi-scale features, the model builds a comprehensive feature set that is robust to variations in signing speed and style, enhancing its ability to interpret novel grammatical compositions.

\subsubsection{Joint Attention Mechanism}
The joint attention mechanism performs a cross-attention operation where the initial input features are used to generate contextual representation. This allows the model to dynamically re-focus on the most salient aspects of the final representation.

Let, \({X}'\) be the initial projected keypoint sequence, and \({H}_{att}\) be the high-level contextual output from the attention weights block. The joint attention mechanism computes its output, \({A}_{joint}\), as shown in (\ref{eq:joint_attention_final}).
\begin{equation}
\label{eq:joint_attention_final}
{A}_{joint} = {F}_{Attention}({X}'{W}^Q, {X}'{W}^K, {H}_{att}{W}^V)
\end{equation}
In this architecture, the Query (Q) and Key (K) are derived from the initial, untransformed features \({X}'\). The Value (V) is derived from the final, context-aware output \({H}_{att}\) of the attention weights block. The intuition is that the model uses raw gestural input to determine which portions of the fully understood sentence are most essential for the final prediction.

\subsubsection{Multi-Scale Dual-Path Encoder}
The challenge of recognizing new sentences requires understanding both fine-grained motions and high-level temporal structure. To this end, we introduce a dual-path temporal encoder that processes the input joint attention module \({A}_{joint}\) in parallel:
\begin{itemize}
\item \textbf{Main Block:} This path uses 1D convolutions to capture fine-grained, frame-level temporal dynamics, preserving the original temporal dimension.
\item \textbf{Auxiliary Block:} This block includes a max-pooling layer with a kernel size and stride of 2 in its convolutional blocks. This downsamples the sequence by a factor of two, enabling the network to learn temporal representations efficiently.
\end{itemize} 
The outputs of these two paths are then concatenated, generating a multi-scale feature set \({F}_{ms} \in \mathbb{R}^{T \times D_{ms}}\) that provides subsequent layers with detailed views of the input sequence. This fusion process is defined as in (\ref{eq:multiscale_fusion}).
\begin{equation}
\label{eq:multiscale_fusion}
{F}_{ms} = F_{concat}(F_{main}({A}_{joint}), F_{aux}({A}_{joint}))
\end{equation}
where \(F_{main}\) is main block and \(F_{aux}\) is auxiliary block from the temporal encoder.

\subsubsection{Transformer Encoder}
The fused multi-scale features, \({F}_{ms}\) are passed into the transformer encoder, which consists of stacked blocks of MHSA and position-wise feed-forward networks. The role of the transformers is to model the long-range dependencies and complex grammatical relationships between the signs in the sequence that is crucial for generalizing to unseen sentence structures.

\subsubsection{Classifier Head}
The contextually-refined representation from the transformer encoder, \({F}_{ms}\), serves as the input to the classifier head. This head consists of a Multi-Layer Perceptron (MLP) with a GELU activation and dropout layers for regularization. This MLP projects the feature dimension of the sequence representation to map the size of the gloss vocabulary, \(|V_g|\). The entire model is trained end-to-end with the CTC loss function, which allows learning without the need for manual frame-by-frame alignment of the video and gloss labels.
\section{Experiments}
\label{sec:experiment} 
This section details the experimental evaluation of our proposed models for CSLR. We describe the dataset, the training and evaluation setup, the baselines used for comparison, and a detailed analysis of the results for both SI and US tasks.

\subsection{Dataset}
All experiments are performed on the Isharah dataset \cite{alyami2025isharah}, a large-scale, multi-scene dataset for Continuous Saudi Sign Language (CSSL) recognition. We used the Isharah-1000 subset for our experiments, as it provides a substantial vocabulary and number of samples while remaining computationally efficient. This subset contains 15,000 videos with 1,000 unique sign-language sentences from 18 signers, a gloss vocabulary of 685, and a text vocabulary of 1,496 words. The main challenge of Isharah is its unconstrained nature, with videos recorded on smartphones in diverse, real-world environments, leading to high variability in lighting, background, and camera angles. We follow the official benchmark splits defined in the original dataset and used in the SignEval 2025 challenge \cite{luqman2025signeval}. 
\begin{itemize}
    \item \textbf{Signer-Independent:} This split evaluates the model's ability to generalize to new signers. The test set consists of videos from four signers who do not appear in the training or development sets.
    \item \textbf{Unseen-Sentences}: This split evaluates generalization to novel sentence structures. The sentences in the development and test sets are unseen during training, while the model has seen the individual glosses in different contexts. This task is challenging due to high singleton and Out-of-Vocabulary (OOV) rates.
\end{itemize}

\subsection{Experimental Setup} The proposed models were implemented using PyTorch. For training, we utilized the AdamW optimizer with a learning rate of \(1 \times 10^{-4}\) and a cosine annealing learning rate scheduler. All models were trained for 100 to 675 epochs with a batch size of 16, distributed across two NVIDIA A6000 GPUs.

\subsection{Evaluation Metric} 
Following the standard protocol for CSLR \cite{alyami2024reviewing} and the Isharah dataset, we measure the WER as our primary evaluation metric. WER is calculated as the sum of Substitutions (S), Insertions (I), and Deletions (D) required to transform the predicted gloss sequence into the ground truth sequence, divided by the number of glosses in the reference (N). The overall calculation process is mentioned in (\ref{eq:wer}). A lower WER indicates better performance.

\begin{equation}
\text{WER} = \frac{S + I + D}{N}
\label{eq:wer}
\end{equation}

\subsection{Baselines}
To evaluate the performance of our proposed models, we compare them against baseline architectures implemented in this research. These models include established and recent approaches in sequence modeling, from classic recurrent networks to hybrid architectures incorporating Large Language Models (LLMs). 
\begin{itemize}
    \item \textbf{LLM-SlowFast:} This model implements the SlowFast \cite{ahn2024slowfast} concept to pose data, with parallel transformer pathways processing the sequence at different temporal dimensions. It further inserts linguistic knowledge by concatenating features from a pretrained XLM-RoBERTa model \cite{conneau2019unsupervised} before the final classifier. 
       
    \item \textbf{LLaMA-Former:} This baseline uses a standard transformer encoder to process pose features, which are then fed into a frozen LLaMA-2 \cite{touvron2023llama} model to act as a sequential processor. This approach explores leveraging the advanced sequence modeling capabilities of a large generative LLM. 

    \item \textbf{LLaMA-SlowFast:} This model fuses LLaMA-2 and a SlowFast architecture to extract multi-rate temporal features from pose data. The fused visual features are then processed by an AraBERT model \cite{antoun2020arabert}. 

    \item \textbf{ST-GCN-Conformer:} This model first employs an ST-GCN to learn features directly on the skeletal graph \cite{yan2018spatial}. The output of the ST-GCN is then processed by a conformer encoder to capture the long-range sequential relationships among these learned spatio-temporal features. 

    \item \textbf{DistilBERT-Former:} This model initially processes the pose sequence using a standard transformer encoder to capture visual-temporal dependencies. The resulting feature embeddings are then fed into a pretrained DistilBERT model \cite{sanh2019distilbert}. This approach aims to leverage the linguistic and contextual knowledge inherent in the LLM backbone. 

    \item \textbf{Mamba-Sign:} A hybrid Mamba-transformer block is utilized in this architecture, replacing traditional attention-based backbones. This design leverages the linear-time sequence modeling strengths of Mamba and the global context capabilities of self-attention \cite{gu2023mamba}. It represents an exploration of recent state-space models for their effectiveness in handling long sequences.
    
    \item \textbf{BiLSTM:} This is a classic CSLR baseline consisting of a simple Bi-directional Long Short-Term Memory (BiLSTM) network \cite{hochreiter1997long}. It processes the pose features directly to capture temporal dependencies.

    \item \textbf{Sign-Conformer:} This network modifies the conformer architecture, which has shown great success in sign language domain. It combines convolutions and self-attention to capture both local and global dependencies in the pose sequence \cite{gulati2020conformer}. 

    \item \textbf{CNN-BiLSTM:} This architecture combines a Temporal Convolutional Network (TCN) \cite{lea2016temporal} with a BiLSTM backbone. The convolutional layers extract and downsample local spatio-temporal features, which are then modeled by the BiLSTM to capture long-range dependencies. 
    
\end{itemize}

\begin{table}[t!]
\centering

\begin{tabular}{l c c}
\toprule
\textbf{Methods} & \textbf{Dev} & \textbf{Test} \\
\midrule
$\text{VAC}^*$ \cite{min2021visual} & 18.9 & 31.9 \\
$\text{SMKD}^*$ \cite{hao2021self} & 18.5 & 35.1 \\
$\text{TLP}^*$ \cite{hu2022temporal} & 19.0 & 32.0 \\
$\text{SEN}^*$ \cite{hu2023self} & 19.1 & 36.4 \\
$\text{CorrNet}^*$ \cite{hu2023continuous} & 18.8 & 31.9 \\
$\text{Swin-MSTP}^*$ \cite{alyami2025swin} & 17.9 & 26.6 \\
$\text{SlowFastSign}^*$ \cite{ahn2024slowfast} & 19.0 & 32.1 \\
\midrule
LLM-SlowFast & 43.90 & 72.24 \\
LLaMA-Former & 21.83 & 51.21 \\
LLaMA-SlowFast & 30.13 & 46.98 \\
Mamba-Sign & 29.31 & 37.28 \\
Multi-Scale Fusion Transformer & 27.54 & 33.91 \\
BiLSTM & 17.02 & 26.08 \\
Sign-Conformer & 16.25 & 26.63 \\
CNN-BiLSTM & \underline{14.54} & \underline{22.62} \\
\midrule
\textbf{Signer-Invariant Conformer} & \textbf{7.31} & \textbf{13.07} \\
\bottomrule
\end{tabular}
\begin{tablenotes}
      \footnotesize
      \item[*] $^{*}\text{results come from \cite {alyami2025isharah}}$.
  \end{tablenotes}
\caption{Performance on the Isharah-1000 SI benchmark. The table compares the WER in percent (\%) on the Development (Dev) and test sets. Lower values indicate better performance. Our proposed models are presented below the SOTA methods and our own implemented baselines. The best result is shown in bold, and the second-best is underlined.}
\label{tab:si_results}
\end{table}

\subsection{Results}
The comparative performance of our proposed architectures against prior works and our implemented baselines is shown in Table \ref{tab:si_results} for SI task and Table \ref{tab:us_results} for US challenge. The following analysis discusses the major findings from these experiments.


\subsubsection{SI Recognition}
On the SI task, our Signer-Invariant Conformer establishes a new SOTA with a test WER of 13.07\% and development WER of 7.31\%. This represents a significant improvement, achieving a relative error reduction of over 50\% compared to the previous best result of 26.6\% from Swin-MSTP \cite{alyami2025swin}. For a direct comparison, the results for these previously published models are taken from the original Isharah benchmark study \cite{alyami2025isharah}. We attribute this substantial gain to the conformer architecture's fusion of convolutional feature extraction for local gestural patterns and self-attention for global temporal context. This combination is highly effective at generating representations invariant to signing style.

It is observed from the experimental findings that our CNN-BiLSTM baseline, achieves a test WER of 22.62\%, outperforming all previously published methods on this dataset. This demonstrates the effectiveness of a robust feature extraction front-end. More complex baselines, such as LLaMA-Former and LLM-SlowFast, exhibited significantly lower outcomes. This shows that complex and domain-specific visual sequence modeling is more important for the recognition task of mapping signals to glosses than relying on generic language models.

\begin{table}[t!]
\centering
\begin{tabular}{l c c}
\toprule
\textbf{Methods} & \textbf{Dev} & \textbf{Test} \\
\midrule
$\text{VAC}^*$ \cite{min2021visual} & 57.0 & \underline{49.6} \\
$\text{SMKD}^*$ \cite{hao2021self} & \underline{56.6} & 48.0 \\
$\text{TLP}^*$ \cite{hu2022temporal} & 70.8 & 63.3 \\
$\text{SEN}^*$ \cite{hu2023self} & 66.2 & 57.3 \\
$\text{CorrNet}^*$ \cite{hu2023continuous} & 63.7 & 55.0 \\
$\text{Swin-MSTP}^*$ \cite{alyami2025swin} & 73.5 & 66.1 \\
$\text{SlowFastSign}^*$ \cite{ahn2024slowfast} & 65.5 & 56.2 \\
\midrule
LLM-SlowFast & 93.07 & - \\
ST-GCN-Conformer & 91.80 & - \\
LLaMA-Former & 86.90 & - \\
DistilBERT-Former & 81.70 & - \\
BiLSTM  & 79.93 & - \\
Sign-Conformer & 77.50 & - \\
CNN-BiLSTM & 74.96 & - \\
Signer-Invariant Conformer & 64.48 & - \\
Mamba-Sign & 59.51 & - \\
\midrule
\textbf{Multi-Scale Fusion Transformer} & \textbf{55.08} & \textbf{47.78} \\
\bottomrule
\end{tabular}
\begin{tablenotes}
      \footnotesize
      \item[*] $^{*}\text{results come from \cite {alyami2025isharah}}$.
  \end{tablenotes}
\caption{Performance on the Isharah-1000 US benchmark. The table compares the WER in percent (\%) on the Development (Dev) and test sets. Lower values indicate better performance. Our proposed models are presented below the SOTA methods and our own implemented baselines. The best result is shown in bold, and the second-best is underlined.}
\label{tab:us_results}
\end{table}

\subsubsection{US Generalization}
The US task is designed to test linguistic generalization, where almost all models exhibiting significantly higher error rates. Our Multi-Scale Fusion Transformer sets a new SOTA with a test and development WER of 47.78\% and 55.08\%. Although the percentage of improvement over the previous best (48.0\% from SMKD \cite{hao2021self}) is low, any improvement on this challenging benchmark is significant. As presented in Table \ref{tab:us_results}, the performance of all prior works is obtained from the official benchmark results in \cite{alyami2025isharah}. The reason behind the enhancement is that the proposed network exploits the joint attention block, temporal encoder and transformer encoder that extract high-level context from the gestural data. This ensures the final representation is contextually-aware to the original pose data, achieving a more robust interpretation of novel sign sequences.

The baseline results further highlight the challenges for this particular task. The architectures that performed well on the SI task, such as CNN-BiLSTM and Sign-Conformer, struggle to generalize, scoring WER above 70\%. The models incorporating ST-GCN and LLM backbones fail in most cases, demonstrating an inability to adapt to unseen syntactic patterns. It is shown that Mamba-Sign emerges as the most competitive among our baselines, suggesting that the state-space models may hold significant features for the linguistic challenges inherent in CSLR.

\section{Conclusion}
\label{sec:conclusion} 
In this work, we introduced a dual-architecture framework to address the distinct issues of SI and US CSLR. Our main contribution is the development of two novel pose-based models: a Signer-Invariant Conformer for signer-agnostic feature learning, and a Multi-Scale Fusion Transformer for linguistic generalization. Our findings on the Isharah-1000 benchmark reveal the efficiency of this approach, with both models achieving new standards. The proposed conformer model significantly reduced absolute WER by 13.53\% on the SI challenge. This supports our proposed solutions and establishes a new benchmark for developing more robust, real-world CSLR systems.

Although efficient, we recognize that our pose-based approach depends on the accuracy of the upstream keypoint extractor, and its efficiency has only been verified on the Isharah dataset. There are many promising ways for future improvement. The next potential step is to apply these advanced encoders to the task of SLT. We also intend to investigate multi-modal fusion, adding RGB features such as hand shape and face expressions to enhance robustness against pose estimation errors. The final research direction involves the development of a unified, multi-task architecture able to handle both SI and US recognition inside a single, efficient framework.

{
    \small
    \bibliographystyle{ieeenat_fullname}
    \bibliography{main}

\begin{thebibliography}{39}
\providecommand{\natexlab}[1]{#1}
\providecommand{\url}[1]{\texttt{#1}}
\expandafter\ifx\csname urlstyle\endcsname\relax
  \providecommand{\doi}[1]{doi: #1}\else
  \providecommand{\doi}{doi: \begingroup \urlstyle{rm}\Url}\fi

\bibitem[who()]{who}
Hearing loss statistics.
\newblock \url{https://www.who.int/newsroom/ fact-sheets/detail/deafness-and-hearing-loss}.
\newblock Accessed: 2025-06-30.

\bibitem[Adaloglou et~al.(2021)Adaloglou, Chatzis, Papastratis, Stergioulas, Papadopoulos, Zacharopoulou, Xydopoulos, Atzakas, Papazachariou, and Daras]{adaloglou2021comprehensive}
Nikolas Adaloglou, Theocharis Chatzis, Ilias Papastratis, Andreas Stergioulas, Georgios~Th Papadopoulos, Vassia Zacharopoulou, George~J Xydopoulos, Klimnis Atzakas, Dimitris Papazachariou, and Petros Daras.
\newblock A comprehensive study on deep learning-based methods for sign language recognition.
\newblock \emph{IEEE Transactions on Multimedia}, 24:\penalty0 1750--1762, 2021.

\bibitem[Ahn et~al.(2024)Ahn, Jang, and Chung]{ahn2024slowfast}
Junseok Ahn, Youngjoon Jang, and Joon~Son Chung.
\newblock Slowfast network for continuous sign language recognition.
\newblock In \emph{ICASSP 2024-2024 IEEE International Conference on Acoustics, Speech and Signal Processing (ICASSP)}, pages 3920--3924. IEEE, 2024.

\bibitem[Aloysius and Geetha(2020)]{aloysius2020understanding}
Neena Aloysius and M Geetha.
\newblock Understanding vision-based continuous sign language recognition.
\newblock \emph{Multimedia Tools and Applications}, 79\penalty0 (31):\penalty0 22177--22209, 2020.

\bibitem[Alyami and Luqman(2025)]{alyami2025swin}
Sarah Alyami and Hamzah Luqman.
\newblock Swin-mstp: Swin transformer with multi-scale temporal perception for continuous sign language recognition.
\newblock \emph{Neurocomputing}, 617:\penalty0 129015, 2025.

\bibitem[Alyami et~al.(2024)Alyami, Luqman, and Hammoudeh]{alyami2024reviewing}
Sarah Alyami, Hamzah Luqman, and Mohammad Hammoudeh.
\newblock Reviewing 25 years of continuous sign language recognition research: Advances, challenges, and prospects.
\newblock \emph{Information Processing \& Management}, 61\penalty0 (5):\penalty0 103774, 2024.

\bibitem[Alyami et~al.(2025)Alyami, Luqman, Al-Azani, Alowaifeer, Alharbi, and Alonaizan]{alyami2025isharah}
Sarah Alyami, Hamzah Luqman, Sadam Al-Azani, Maad Alowaifeer, Yazeed Alharbi, and Yaser Alonaizan.
\newblock Isharah: A large-scale multi-scene dataset for continuous sign language recognition.
\newblock \emph{arXiv:2506.03615}, 2025.

\bibitem[Antoun et~al.(2020)Antoun, Baly, and Hajj]{antoun2020arabert}
Wissam Antoun, Fady Baly, and Hazem Hajj.
\newblock Arabert: Transformer-based model for arabic language understanding.
\newblock \emph{arXiv:2003.00104}, 2020.

\bibitem[Athira et~al.(2022)Athira, Sruthi, and Lijiya]{athira2022signer}
PK Athira, CJ Sruthi, and A Lijiya.
\newblock A signer independent sign language recognition with co-articulation elimination from live videos: an indian scenario.
\newblock \emph{Journal of King Saud University-Computer and Information Sciences}, 34\penalty0 (3):\penalty0 771--781, 2022.

\bibitem[Cihan~Camgoz et~al.(2017)Cihan~Camgoz, Hadfield, Koller, and Bowden]{cihan2017subunets}
Necati Cihan~Camgoz, Simon Hadfield, Oscar Koller, and Richard Bowden.
\newblock Subunets: End-to-end hand shape and continuous sign language recognition.
\newblock In \emph{Proceedings of the IEEE International Conference on Computer Vision}, pages 3056--3065, 2017.

\bibitem[Conneau et~al.(2019)Conneau, Khandelwal, Goyal, Chaudhary, Wenzek, Guzm{\'a}n, Grave, Ott, Zettlemoyer, and Stoyanov]{conneau2019unsupervised}
Alexis Conneau, Kartikay Khandelwal, Naman Goyal, Vishrav Chaudhary, Guillaume Wenzek, Francisco Guzm{\'a}n, Edouard Grave, Myle Ott, Luke Zettlemoyer, and Veselin Stoyanov.
\newblock Unsupervised cross-lingual representation learning at scale.
\newblock \emph{arXiv:1911.02116}, 2019.

\bibitem[Cui et~al.(2019)Cui, Liu, and Zhang]{cui2019deep}
Runpeng Cui, Hu Liu, and Changshui Zhang.
\newblock A deep neural framework for continuous sign language recognition by iterative training.
\newblock \emph{IEEE Transactions on Multimedia}, 21\penalty0 (7):\penalty0 1880--1891, 2019.

\bibitem[Cui et~al.(2023)Cui, Zhang, Li, and Wang]{cui2023spatial}
Zhenchao Cui, Wenbo Zhang, Zhaoxin Li, and Zhaoqi Wang.
\newblock Spatial--temporal transformer for end-to-end sign language recognition.
\newblock \emph{Complex \& Intelligent Systems}, 9\penalty0 (4):\penalty0 4645--4656, 2023.

\bibitem[De~Sisto et~al.(2023)De~Sisto, Sequino, and Ciaramella]{desisto2023survey}
Maria De~Sisto, Gaetano Sequino, and Angelo Ciaramella.
\newblock A survey on continuous sign language recognition.
\newblock \emph{ACM Computing Surveys}, 55\penalty0 (10):\penalty0 1--38, 2023.

\bibitem[El-Alfy and Luqman(2022)]{el2022comprehensive}
El-Sayed~M El-Alfy and Hamzah Luqman.
\newblock A comprehensive survey and taxonomy of sign language research.
\newblock \emph{Engineering Applications of Artificial Intelligence}, 114:\penalty0 105198, 2022.

\bibitem[Forster et~al.(2014)Forster, Schmidt, Koller, Bellgardt, and Ney]{forster2014extensions}
Jens Forster, Christoph Schmidt, Oscar Koller, Martin Bellgardt, and Hermann Ney.
\newblock Extensions of the sign language recognition and translation corpus rwth-phoenix-weather.
\newblock In \emph{LREC}, pages 1911--1916, 2014.

\bibitem[Gu and Dao(2023)]{gu2023mamba}
Albert Gu and Tri Dao.
\newblock Mamba: Linear-time sequence modeling with selective state spaces.
\newblock \emph{arXiv:2312.00752}, 2023.

\bibitem[Gulati et~al.(2020)Gulati, Qin, Chiu, Parmar, Zhang, Yu, Han, Wang, Zhang, Wu, et~al.]{gulati2020conformer}
Anmol Gulati, James Qin, Chung-Cheng Chiu, Niki Parmar, Yu Zhang, Jiahui Yu, Wei Han, Shibo Wang, Zhengdong Zhang, Yonghui Wu, et~al.
\newblock Conformer: Convolution-augmented transformer for speech recognition.
\newblock \emph{arXiv:2005.08100}, 2020.

\bibitem[Hao et~al.(2021)Hao, Min, and Chen]{hao2021self}
Aiming Hao, Yuecong Min, and Xilin Chen.
\newblock Self-mutual distillation learning for continuous sign language recognition.
\newblock In \emph{Proceedings of the IEEE/CVF International Conference on Computer Vision}, pages 11303--11312, 2021.

\bibitem[Hochreiter and Schmidhuber(1997)]{hochreiter1997long}
Sepp Hochreiter and J{\"u}rgen Schmidhuber.
\newblock Long short-term memory.
\newblock \emph{Neural Computation}, 9\penalty0 (8):\penalty0 1735--1780, 1997.

\bibitem[Hu et~al.(2022)Hu, Gao, Liu, and Feng]{hu2022temporal}
Lianyu Hu, Liqing Gao, Zekang Liu, and Wei Feng.
\newblock Temporal lift pooling for continuous sign language recognition.
\newblock In \emph{European Conference on Computer Vision}, pages 511--527. Springer, 2022.

\bibitem[Hu et~al.(2023{\natexlab{a}})Hu, Gao, Liu, and Feng]{hu2023continuous}
Lianyu Hu, Liqing Gao, Zekang Liu, and Wei Feng.
\newblock Continuous sign language recognition with correlation network.
\newblock In \emph{Proceedings of the IEEE/CVF Conference on Computer Vision and Pattern Recognition}, pages 2529--2539, 2023{\natexlab{a}}.

\bibitem[Hu et~al.(2023{\natexlab{b}})Hu, Gao, Liu, and Feng]{hu2023self}
Lianyu Hu, Liqing Gao, Zekang Liu, and Wei Feng.
\newblock Self-emphasizing network for continuous sign language recognition.
\newblock In \emph{Proceedings of the AAAI Conference on Artificial Intelligence}, pages 854--862, 2023{\natexlab{b}}.

\bibitem[Huang et~al.(2018)Huang, Zhou, Zhang, Li, and Li]{huang2018video}
Jie Huang, Wengang Zhou, Qilin Zhang, Houqiang Li, and Weiping Li.
\newblock Video-based sign language recognition without temporal segmentation.
\newblock In \emph{Proceedings of the AAAI Conference on Artificial Intelligence}, 2018.

\bibitem[Kagirov et~al.(2020)Kagirov, Ivanko, Ryumin, Axyonov, and Karpov]{kagirov2020theruslan}
Ildar Kagirov, Denis Ivanko, Dmitry Ryumin, Alexander Axyonov, and Alexey Karpov.
\newblock Theruslan: Database of russian sign language.
\newblock In \emph{Proceedings of the Twelfth Language Resources and Evaluation Conference}, pages 6079--6085, 2020.

\bibitem[Lea et~al.(2016)Lea, Vidal, Reiter, and Hager]{lea2016temporal}
Colin Lea, Rene Vidal, Austin Reiter, and Gregory~D Hager.
\newblock Temporal convolutional networks: A unified approach to action segmentation.
\newblock In \emph{Computer vision--ECCV 2016 workshops: Amsterdam, the Netherlands, October 8-10 and 15-16, 2016, proceedings, part III 14}, pages 47--54. Springer, 2016.

\bibitem[Li et~al.(2020)Li, Rodriguez, Yu, and Li]{li2020word}
Dongxu Li, Cristian Rodriguez, Xin Yu, and Hongdong Li.
\newblock Word-level deep sign language recognition from video: A new large-scale dataset and methods comparison.
\newblock In \emph{Proceedings of the IEEE/CVF Winter Conference on Applications of Computer Vision}, pages 1459--1469, 2020.

\bibitem[Li and Meng(2022)]{li2022multi}
Ronghui Li and Lu Meng.
\newblock Multi-view spatial-temporal network for continuous sign language recognition.
\newblock \emph{arXiv:2204.08747}, 2022.

\bibitem[Luqman and Mahmoud(2019)]{luqman2019automatic}
Hamzah Luqman and Sabri~A Mahmoud.
\newblock Automatic translation of arabic text-to-arabic sign language.
\newblock \emph{Universal Access in the Information Society}, 18\penalty0 (4):\penalty0 939--951, 2019.

\bibitem[Luqman et~al.(2025)Luqman, Mineo, Aljubran, Hasanaath, Sorrenti, Alyami, Al-Azani, Alowaifeer, Moon, Javorek, Zelezny, Hruz, Caligiore, Giancola, Polikovsky, Alfarraj, Fontana, Mahmud, Khan, Islam, Gurbuz, Ragonese, Bellitto, Proietto~Salanitri, Spampinato, and Palazzo]{luqman2025signeval}
Hamzah Luqman, Raffaele Mineo, Murtadha Aljubran, Ahmed~Abul Hasanaath, Amelia Sorrenti, Sarah Alyami, Sadam Al-Azani, Maad Alowaifeer, JiHwan Moon, Vaclav Javorek, Tomas Zelezny, Marek Hruz, Gaia Caligiore, Silvio Giancola, Senya Polikovsky, Motaz Alfarraj, Sabina Fontana, Mufti Mahmud, Muhammad~Haris Khan, Kamrul Islam, Sevgi Gurbuz, Egidio Ragonese, Giovanni Bellitto, Federica Proietto~Salanitri, Concetto Spampinato, and Simone Palazzo.
\newblock The signeval 2025 challenge at the iccv multimodal sign language recognition workshop: Results and discussion.
\newblock In \emph{2025 IEEE/CVF International Conference on Computer Vision Workshops}, 2025.

\bibitem[Min et~al.(2021)Min, Hao, Chai, and Chen]{min2021visual}
Yuecong Min, Aiming Hao, Xiujuan Chai, and Xilin Chen.
\newblock Visual alignment constraint for continuous sign language recognition.
\newblock In \emph{proceedings of the IEEE/CVF International Conference on Computer Vision}, pages 11542--11551, 2021.

\bibitem[Mukushev et~al.(2022)Mukushev, Ubingazhibov, Kydyrbekova, Imashev, Kimmelman, and Sandygulova]{mukushev2022fluentsigners}
Medet Mukushev, Aidyn Ubingazhibov, Aigerim Kydyrbekova, Alfarabi Imashev, Vadim Kimmelman, and Anara Sandygulova.
\newblock Fluentsigners-50: A signer independent benchmark dataset for sign language processing.
\newblock \emph{PLOS One}, 17\penalty0 (9):\penalty0 e0273649, 2022.

\bibitem[Sanh et~al.(2019)Sanh, Debut, Chaumond, and Wolf]{sanh2019distilbert}
Victor Sanh, Lysandre Debut, Julien Chaumond, and Thomas Wolf.
\newblock Distilbert, a distilled version of bert: smaller, faster, cheaper and lighter.
\newblock \emph{arXiv:1910.01108}, 2019.

\bibitem[Sidig et~al.(2021)Sidig, Luqman, Mahmoud, and Mohandes]{sidig2021karsl}
Ala Addin~I Sidig, Hamzah Luqman, Sabri Mahmoud, and Mohamed Mohandes.
\newblock Karsl: Arabic sign language database.
\newblock \emph{ACM Transactions on Asian and Low-Resource Language Information Processing (TALLIP)}, 20\penalty0 (1):\penalty0 1--19, 2021.

\bibitem[Touvron et~al.(2023)Touvron, Martin, Stone, Albert, Almahairi, Babaei, Bashlykov, Batra, Bhargava, Bhosale, et~al.]{touvron2023llama}
Hugo Touvron, Louis Martin, Kevin Stone, Peter Albert, Amjad Almahairi, Yasmine Babaei, Nikolay Bashlykov, Soumya Batra, Prajjwal Bhargava, Shruti Bhosale, et~al.
\newblock Llama 2: Open foundation and fine-tuned chat models.
\newblock \emph{arXiv preprint arXiv:2307.09288}, 2023.

\bibitem[von Agris and Kraiss(2010)]{von2010signum}
Ulrich von Agris and Karl-Friedrich Kraiss.
\newblock Signum database: Video corpus for signer-independent continuous sign language recognition.
\newblock In \emph{4th Workshop on the Representation and Processing of Sign Languages: Corpora and Sign Language Technologies}, pages 243--246, 2010.

\bibitem[Yan et~al.(2018)Yan, Xiong, and Lin]{yan2018spatial}
Sijie Yan, Yuanjun Xiong, and Dahua Lin.
\newblock Spatial temporal graph convolutional networks for skeleton-based action recognition.
\newblock In \emph{Proceedings of the AAAI Conference on Artificial Intelligence}, 2018.

\bibitem[Zhang et~al.(2020)Zhang, Zhang, and Zheng]{10.1145/3377553}
Lei Zhang, Yixiang Zhang, and Xiaolong Zheng.
\newblock Wisign: Ubiquitous american sign language recognition using commercial wi-fi devices.
\newblock \emph{ACM Transactions on Intelligent Systems and Technology}, 11\penalty0 (3), 2020.

\bibitem[Zhou et~al.(2020)Zhou, Zhou, Zhou, and Li]{zhou2020spatial}
Hao Zhou, Wengang Zhou, Yun Zhou, and Houqiang Li.
\newblock Spatial-temporal multi-cue network for continuous sign language recognition.
\newblock In \emph{Proceedings of the AAAI Conference on Artificial Intelligence}, pages 13009--13016, 2020.

\end{thebibliography}
}

\end{document}